\documentclass{article}

\usepackage{microtype}
\usepackage{graphicx}
\usepackage{subfigure}
\usepackage{booktabs} % for professional tables

\usepackage{hyperref}
\usepackage{multirow}
\usepackage{enumitem}

\usepackage{nccmath}  
\usepackage{bbm}
\usepackage{amsmath}
\usepackage{amssymb}
\usepackage{mathtools}
\usepackage{amsthm}

\usepackage{amsfonts}
\usepackage{gensymb}

\usepackage{color, colortbl}
\usepackage{skak}

\usepackage[accepted]{icml2023}

\usepackage{amsmath}
\usepackage{amssymb}
\usepackage{mathtools}
\usepackage{amsthm}
\usepackage{xcolor}

\usepackage[capitalize,noabbrev]{cleveref}

\definecolor{highlight}{rgb}{1, 0.94, 0.87}

\newcommand{\tf}[1]{\mathbf{#1}}
\newcommand{\tfg}[1]{\boldsymbol{#1}}
\newcommand{\R}{\mathbb{R}}

\newcommand{\cmark}{\ding{51}}

\newcommand*\dif{\mathop{}\!\mathrm{d}}

\usepackage{pifont}
\newcommand\mynuma[1]{\ifcase#1 \or \ding{172}\or \ding{173}\or
  \ding{174}\or \ding{175}\or \ding{176}\or \ding{177}%
  \or \ding{178}\or \ding{179}\or \ding{180}\or \ding{181}\else *\fi\relax}

\newcommand\mynumb[1]{\ifcase#1 \or \ding{182}\or \ding{183}\or
  \ding{184}\or \ding{185}\or \ding{186}\or \ding{187}%
  \or \ding{188}\or \ding{189}\or \ding{190}\or \ding{191}\else *\fi\relax}

\definecolor{gray}{gray}{0.9}

\theoremstyle{plain}

\theoremstyle{definition}

\theoremstyle{remark}

\usepackage[textsize=tiny]{todonotes}

\icmltitlerunning{NeRFool: Uncovering the Vulnerability of Generalizable Neural Radiance Fields against Adversarial Perturbations} 
\begin{document}

\twocolumn[
\icmltitle{NeRFool: Uncovering the Vulnerability of Generalizable Neural \\ Radiance Fields against Adversarial Perturbations}

\begin{icmlauthorlist}
\icmlauthor{Yonggan Fu}{yyy}
\icmlauthor{Ye Yuan}{yyy}
\icmlauthor{Souvik Kundu}{comp}
\icmlauthor{Shang Wu}{yyy}
\icmlauthor{Shunyao Zhang}{xxx}
\icmlauthor{Yingyan (Celine) Lin}{yyy}
\end{icmlauthorlist}

\icmlaffiliation{yyy}{School of Computer Science, Georgia Institute of Technology, USA}
\icmlaffiliation{comp}{Intel Labs, San Diego, USA}
\icmlaffiliation{xxx}{Rice University, USA}

\icmlcorrespondingauthor{Yingyan (Celine) Lin}{celine.lin@gatech.edu}

% You may provide any keywords that you
% find helpful for describing your paper; these are used to populate
% the "keywords" metadata in the PDF but will not be shown in the document
\icmlkeywords{Machine Learning, ICML}

\vskip 0.3in
]

% this must go after the closing bracket ] following \twocolumn[ ...

% This command actually creates the footnote in the first column
% listing the affiliations and the copyright notice.
% The command takes one argument, which is text to display at the start of the footnote.
% The \icmlEqualContribution command is standard text for equal contribution.
% Remove it (just {}) if you do not need this facility.

\printAffiliationsAndNotice{}  % leave blank if no need to mention equal contribution
%\printAffiliationsAndNotice{\icmlEqualContribution} % otherwise use the standard text.

\begin{abstract}

Generalizable Neural Radiance Fields (GNeRF) are one of the most promising real-world solutions for novel view synthesis, thanks to their cross-scene generalization capability and thus the possibility of instant rendering on new scenes. While adversarial robustness is essential for real-world applications, little study has been devoted to understanding its implication on GNeRF. We hypothesize that because GNeRF is implemented by conditioning on the source views from new scenes, which are often acquired from the Internet or third-party providers, there are potential new security concerns regarding its real-world applications. Meanwhile, existing understanding and solutions for neural networks' adversarial robustness may not be applicable to GNeRF, due to its 3D nature and uniquely diverse operations. To this end, we present NeRFool, which to the best of our knowledge is the first work that sets out to understand the adversarial robustness of GNeRF. Specifically, NeRFool unveils the vulnerability patterns and important insights regarding GNeRF's adversarial robustness. Built upon the above insights gained from NeRFool, we further develop NeRFool$^+$, which integrates two techniques capable of effectively attacking GNeRF across a wide range of target views, and provide guidelines for defending against our proposed attacks. We believe that our NeRFool/NeRFool$^+$ lays the initial foundation for future innovations in developing robust real-world GNeRF solutions.
Our codes are available at: \href{https://github.com/GATECH-EIC/NeRFool}{https://github.com/GATECH-EIC/NeRFool}.

\end{abstract}

% \vspace{-1em}
\section{Introduction}
\label{sec:intro}
 % \vspace{-0.3em}

Novel view synthesis (NVS), which aims to generate photorealistic novel views of a scene given only a set of sparsely sampled views, has become an essential functionality in real-world 3D vision applications. Among various NVS techniques, neural radiance fields (NeRF)~\cite{mildenhall2021nerf} have recently gained substantial attention thanks to their record-breaking rendering quality, igniting a tremendous demand for NeRF-based NVS solutions. As many real-world NVS applications require instant and real-time rendering on new scenes, generalizable NeRF (GNeRF) variants~\cite{yu2021pixelnerf,wang2021ibrnet,chen2021mvsnerf, liu2022neural} have emerged as  the most appealing real-world NeRF solutions. In particular, GNeRF conditions NeRF on the source views from a new target scene to achieve cross-scene generalization and enable new scene reconstruction via only a single forward pass execution.

Despite GNeRF's big promise towards real-world NVS solutions, it is currently unclear whether it can fulfill the essential robustness requirements. In fact, we hypothesize that GNeRF's introduced conditionality on source views can cause new security concerns. This is because the source views that describe a new scene, e.g., a tourist attraction, are often acquired from the Internet/third-party providers, leaving opportunities for adversaries to take advantage in terms of malicious attacks. 
For example, adversarial perturbations~\cite{goodfellow2014explaining, madry2017towards} can be injected into source views by adversaries to severely degrade the reconstruction accuracy of GNeRF. With the increasing deployment of NeRF-powered security-critical applications, such as robot navigation systems~\cite{adamkiewicz2022vision, maggio2022loc, moreau2022lens} and autonomous driving systems~\cite{kundu2022panoptic, fu2022panoptic, siddiqui2023panoptic}, it is imperative to understand the adversarial robustness of GNeRF for the unleashing of its cross-scene generalization capability toward real-world NeRF-based NVS solutions.

To address the imperative need above, one may naturally consider borrowing the existing insights about the adversarial robustness of deep neural networks (DNNs). However, those insights may not be applicable to NeRF due to its unique properties and processing pipeline. \underline{First}, unlike 2D tasks, the 3D nature of NVS tasks requires NeRF to reconstruct the target 3D scenes across different views. As such, it is not straightforward how to ensure that perturbing 2D source views of a scene can effectively 
pollute the entire 3D scene. \underline{Second}, NeRF features a volume rendering process, in which pixels are rendered via alpha compositing~\cite{mildenhall2021nerf} from estimated density and color, and thus involves more diverse operations than DNNs. Hence, it is unclear which component (e.g., the density/color or both) of NeRF is more vulnerable (or needs stronger protection). \underline{Third}, the ray marching process of GNeRF relies on the geometric relationship among different views, and thus perturbations optimized for destructing one view may be effective for destructing another view. This poses new risks of adversarial perturbations targeting GNeRF which could be transferable across a wide range of views.

To this end, this work sets out to \underline{(1)} raise the community's awareness regarding the potential security concerns of GNeRF due to adversarial perturbations and \underline{(2)} enhance our understanding of GNeRF's vulnerability patterns. We summarize our contributions as follows:

  \vspace{-0.5em}
\begin{itemize}
    \item  We present both NeRFool and NeRFool$^+$, which to the best of our knowledge are \textbf{the first} works that uncover and study the vulnerability of GNeRF against adversarial perturbations. As such, NeRFFool/NeRFool$^+$ open up a new perspective in NeRF literature and can shed light on future innovations toward robust real-world 
    GNeRF-based NVS solutions. 
    
    \item In NeRFool, we study the vulnerability patterns of various GNeRF variants through systematic analysis and experiments, and discover that, interestingly, \underline{(1)} increased conditionality on source views can cause a higher vulnerability of GNeRF and \underline{(2)} adversarial perturbations on density have a significantly stronger ``ruining" ability than that on color when attacking GNeRF, especially on scenes with complex geometry.

    \item  Built upon the above insights gained from NeRFool, we further develop NeRFool$^+$, which integrates two optimization techniques, novel target view sampling and  geometric error maximization, that can 
    effectively attack GNeRF across a wide range of target views.

    \item We further embark on an intriguing exploration to defend against our NeRFool attacks and discover the benign impact of adversarial perturbations on GNeRF's reconstruction accuracy, deepening the understanding regarding GNeRF's robustness.

\end{itemize}

\section{Related Works}
 % \vspace{-0.3em}
\label{sec:related-work}

\textbf{View synthesis and NeRF.}
View synthesis renders photorealistic images from novel views of a scene based on a set of sparsely sampled views~\cite{hedman2018,thies2019neural,lombardi19,mildenhall2021nerf}. Among existing techniques, 
NeRF~\cite{mildenhall2021nerf}, which implicitly represents a scene as a continuous 5D radiance field parameterized by a multilayer perceptron (MLP), has gained increasing popularity.
Follow-up works \underline{(1)} improve NeRF's rendering quality under extremely sparse views~\cite{xu2022sinnerf,niemeyer2022regnerf}, \underline{(2)} accelerate NeRF via reducing the complexity of MLP~\cite{lindell2021autoint,rebain21} or exploring the free space via 3D occupancy grids~\cite{yu2021plenoctrees,garbin2021fastnerf}, and \underline{(3)} extend NeRF to other tasks, e.g., generative modeling~\cite{chan2020pi, schwarz2020graf}, dynamic scenes~\cite{li2021, ost2020neural}, or lighting/reflection modeling~\cite{nerv2021, verbin2022ref}.

\textbf{Generalizable NeRFs.} To avoid tedious per-scene optimization and endow NeRF with cross-scene generalization capability, generalizable NeRFs~\cite{yu2021pixelnerf,wang2021ibrnet} have been developed to reconstruct the radiance field of a new scene via merely a one-shot forward pass.
Specifically, recent GNeRF techniques~\cite{yu2021pixelnerf,wang2021ibrnet,reizenstein2021common,wang2022attention,chen2021mvsnerf,xu2022point,liu2022neural} are implemented by conditioning vanilla NeRF techniques on a set of source views from the new scene via taking the extracted scene features from the source views as inputs. 
Despite their promise, GNeRF's pipeline, i.e., conditioning NeRF on source views, leaves opportunities for adversaries to take advantage in terms of malicious attacks. 
For example, adversaries can attack GNeRF by injecting adversarial perturbations onto the aforementioned source views. 
Hence, it is crucial to understand GNeRF's adversarial robustness for ensuring their real-world deployment, which has yet to be explored by the literature.

 \vspace{-0.3em}
\textbf{Adversarial attack and defense.} DNNs are well-recognized to be adversarially vulnerable~\cite{goodfellow2014explaining}. 
Various attacks~\cite{madry2017towards, carlini2017towards, andriushchenko2020square} are proposed to aggressively degrade the achievable accuracy of DNNs
for different tasks~\cite{arnab2018robustness,carlini2018audio,zhang2020adversarial}. 
In parallel, a variety of defense schemes~\cite{guo2017countering, feinman2017detecting,madry2017towards,shafahi2019adversarial,wong2019fast} is developed to enhance DNNs' adversarial robustness. 
The readers are referred to~\cite{akhtar2018threat, chakraborty2018adversarial} for more attack and defense methods. Nevertheless, 
to the best of our knowledge, \textbf{no existing work has been dedicated to studying NeRF's adversarial robustness.} Instead, recent works have attempted to combine adversarial optimization with NeRF for various purposes. For example,~\cite{chen2022aug} improves NeRF's accuracy using augmented data obtained from perturbing the input coordinates or intermediate features,~\cite{niemeyer2021giraffe,wang2022clip} incorporate adversarial objectives to enhance the reconstruction quality, and~\cite{dong2022viewfool} identifies adversarial viewpoints from which the rendered images can fool downstream image classifiers, instead of aiming to degrade NeRF's own accuracy. 
One concurrent work~\cite{wang2023benchmarking} provides an investigation of NeRF's robustness to common image corruptions but adversarial perturbations are not considered. 
With the growing demand for real-world NeRF-based NVS solutions, it is imperative to understand NeRF's adversarial robustness.

 % \vspace{-0.5em}
\section{Preliminaries of NeRF and GNeRF}
\label{sec:preliminaries}
 % \vspace{-0.3em}

\textbf{NeRF's rendering pipeline.} In NeRF, each 2D pixel on the image plane corresponds to a camera ray $\tf r(t) = \tf o + t \tf d$  emitted from the camera center  $\tf o \in \R^3$, with $\tf d \in \R^3$ denoting the ray direction and $t$ denoting the ray depth. To render a pixel, a NeRF function $f$ samples points along the corresponding ray and then acquires the color $\tf c$ and density $\sigma$ of each point, i.e.,  \((\sigma, \tf c)\ = f(\tf r (t), \tf d)\). 
Next, the 2D pixel $\hat{\mathbf{C}}(\mathbf{r})$ can be derived via an integral over the colors of the above-sampled points: 

\vspace{-0.5em}
\begin{equation}
     \hat{\mathbf{C}}(\mathbf{r}, f) = \int_{t_n}^{t_f} T(t) \sigma(\tf r(t)) \mathbf{c}(\tf r(t), \tf d) \dif t  %\\
 \label{eq:rendering}
\end{equation}
\vspace{-1em}

\noindent where $t_n$ and $t_f$ are the predefined near and far bounds, respectively, and $T(t) = \exp\left(- \int_{t_n}^{t} \sigma(\tf r(s)) \,\dif s \right)$ denotes the accumulated transmittance along the ray from $t_n$ to $t$. In practice, the integral in Eq.~(\ref{eq:rendering}) is often approximated with numerical quadrature~\cite{mildenhall2021nerf}. Finally, an MSE loss is applied between the rendered pixels $\hat{\mathbf{C}}(\mathbf{r}, f)$ and the ground truth pixels $\mathbf{C}(\mathbf{r})$ to train NeRF $f$:

\vspace{-0.5em}
\begin{equation}
    \mathcal{L}_{rgb}(\mathcal{R}, f) = \sum_{\tf r \in \mathcal{R}} \left\lVert
   \hat{\mathbf{C}}(\tf r, f) - \mathbf{C}(\tf r) \right\rVert_2^2
    \label{eq:mseloss}
\end{equation}
\vspace{-1.5em}

\noindent where $\mathcal{R}$ is the set of sampled camera rays.

\textbf{GNeRF's pipeline.} On top of vanilla NeRF's pipeline above, GNeRF conditions its function $f$ on the source views as priors of the target new scenes to enable cross-scene generalization. 
For example,~\cite{wang2021ibrnet, yu2021pixelnerf, wang2022attention, liu2022neural,chen2021mvsnerf} adopt a CNN encoder $E:\mathbb{R}^3\rightarrow\mathbb{R}^3$ to extract features $\{E(\tf I_i)\}$ from the source views $\{\tf I_i\}$. 
Then, each sampled point $\tf x$ on ray $\tf r (t)$ is projected to the image plane of each source view through a transformation $\pi_i:\mathbb{R}^3\rightarrow\mathbb{R}^2$ to acquire the corresponding scene feature $E(\tf I_i)[\pi_i(\tf x)]$. Finally, the acquired features $\tf e=\{E(\tf I_i)[\pi_i(\tf x)]\}$ serve as extra inputs of vanilla NeRF to derive the density and color $(\sigma, \tf c) = f(\tf x, \tf d, \tf e)$. Different GNeRF variants differ in the ways of constructing the above scene features while following both the volume rendering and the objective in Eq.~(\ref{eq:rendering}) and Eq.~(\ref{eq:mseloss}), respectively. As introduced in Sec.~\ref{sec:exploration} and Sec.~\ref{sec:nerfool}, our NeRFool adversarially perturbs the source views $\{\tf I_i\}$, thereby inducing adversarial features in $\tf e$.   
\begin{figure}[t]

% \vspace{-1em}
\centering
\includegraphics[width=0.98\linewidth]{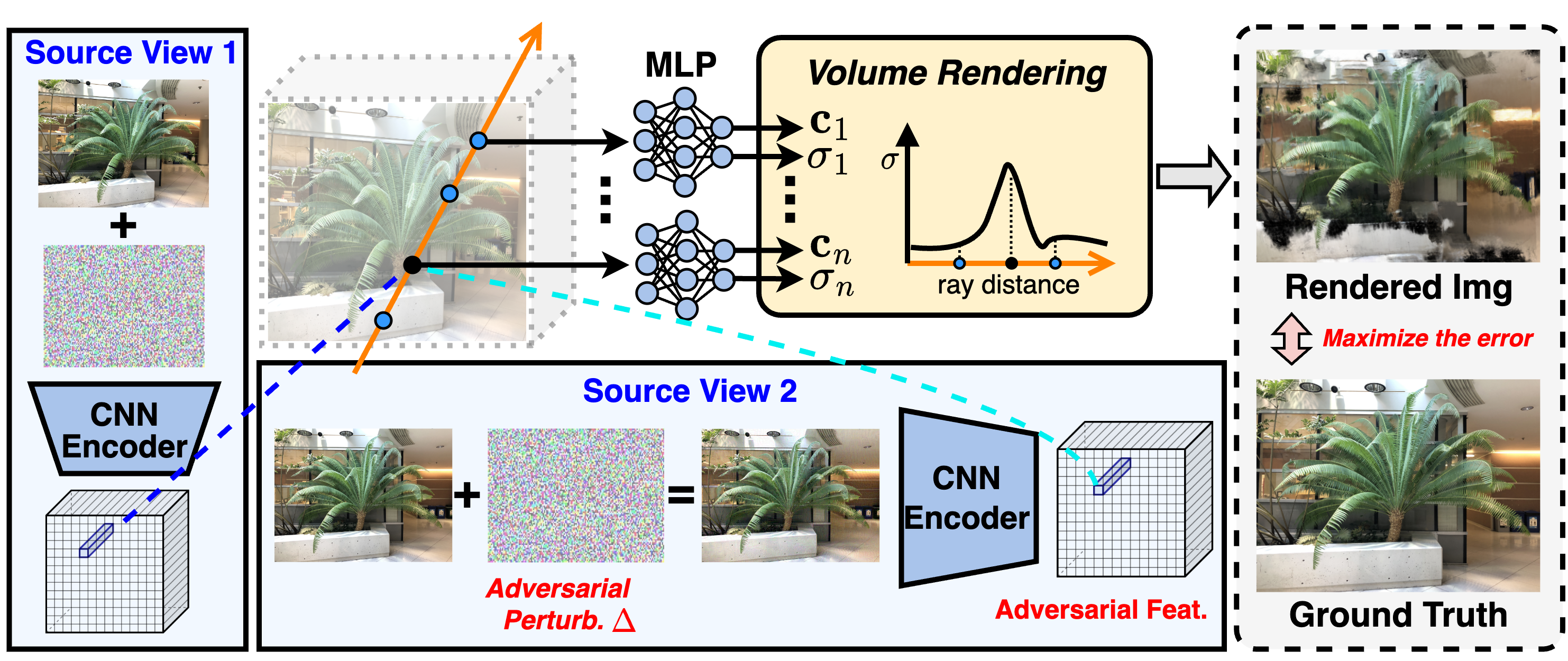}
\vspace{-1em}
\caption{An overview of our NeRFool framework.}
\label{fig:overview}
\vspace{-1em}
\end{figure}

 \begin{figure*}[t]
% \vspace{-0.3em}
\centering
\includegraphics[width=0.98\linewidth]{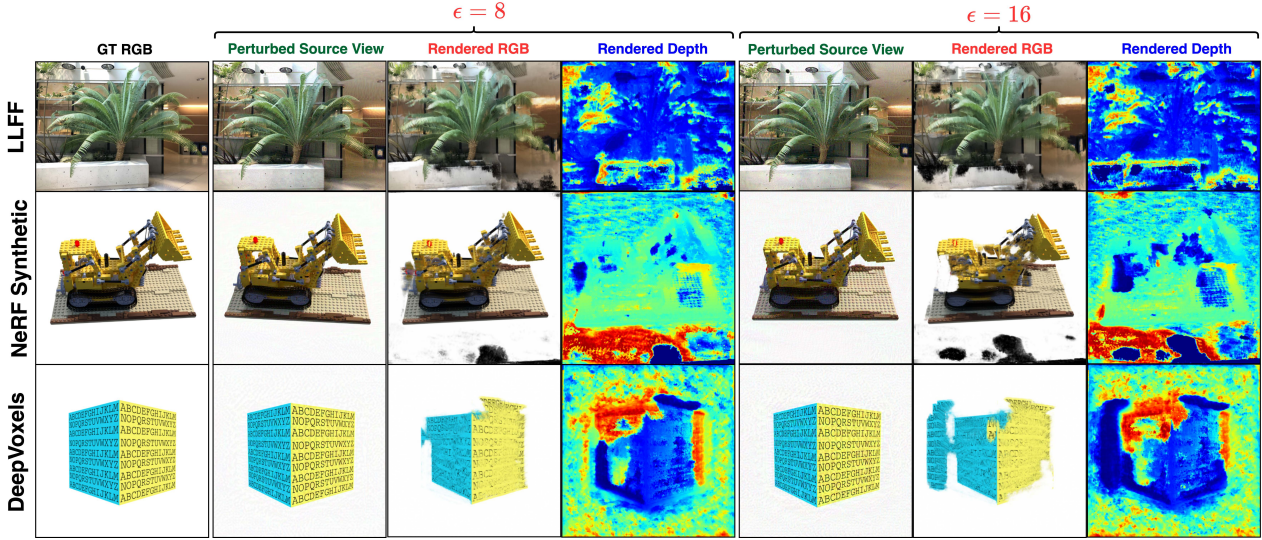}
\vspace{-0.5em}
\caption{Visualize the ground-truth RGB images, adversarially perturbed source views with imperceptible perturbations which are used to attack IBRNet, and the resulting rendered RGB images/depths on three scenes from three datasets.}
\label{fig:ibrnet_vc}
\vspace{-1em}
\end{figure*}

% \vspace{-0.5em} 
\section{NeRFool: Uncover GNeRF's Vulnerability}  
\label{sec:exploration}
 % \vspace{-0.3em}
 
In this section, we present NeRFool, which studies the important properties of GNeRF's vulnerability via our proposed view-specific attack scheme, in which the adversarial perturbations are optimized to fool one specific target view. The insights drawn from NeRFool further inspire our development of NeRFool$^+$ introduced in Sec.~\ref{sec:nerfool}.

\subsection{A View-Specific Method for Attacking GNeRF} 
\label{sec:nerfool-}
 % \vspace{-0.3em}

\textbf{Formulation.}  
As visualized in Fig.~\ref{fig:overview}, we inject adversarial perturbations $\tfg\Delta=\{\tfg \delta_i\}$ into the source view images $\{\tf I_i\}$ correspondingly to degrade the reconstruction accuracy of a GNeRF model on one specific target view with a camera pose $\tf P_{tar}=[\tf R_{tar} \mid \tf T_{tar} ]\in\mathbb{R}^{3\times4}$, where $\tf R_{tar} \in \R^{3\times3}$ and $\tf T_{tar} \in \R^3$ are the rotation and translation with respect to the world coordinate. To optimize $\tfg\Delta$, the goal is to maximize the reconstruction error under a norm constraint $\| \tfg \delta_i\|_{\infty} \leq \epsilon$, 
with $\epsilon$ being sufficiently small to ensure the perturbation's imperceptibility to human eyes.  Specifically, the objective can be formulated as:

\vspace{-0.5em}
\begin{equation} \label{eq:objective}
\max_{\forall \tfg \delta_i \in \tfg\Delta: \, \| \tfg \delta_i\|_{\infty} \leq \epsilon} \tilde{\mathcal{L}}_{rgb}(\mathcal{R}_{tar}, f, \tfg\Delta)
\end{equation} 
\vspace{-1.5em}

where $\mathcal{R}_{tar}$ is the set of rays sampled from the target view $\tf P_{tar}$. Although $\mathcal{L}_{rgb}$ in Eq.~(\ref{eq:mseloss}) can provide effective supervision for optimizing $\tfg\Delta$, the ground truth $\mathbf{C}(\tf r)$ for calculating $\mathcal{L}_{rgb}$ may not be available for all target views. Fortunately, we find that the corresponding pseudo ground truth can be obtained in GNeRF by reconstructing the specified target view based on the clean source views. Accordingly, we can modify $\mathcal{L}_{rgb}$ to $\tilde{\mathcal{L}}_{rgb}$ as our objective function, as formulated below:

\vspace{-0.5em}
\begin{equation} 
\label{eq:rgb_new}
\begin{aligned} 
    \hspace{-1em} \tilde{\mathcal{L}}_{rgb}(\mathcal{R}, f, \tfg\Delta) = \sum_{\tf r \in \mathcal{R}} \big\lVert \hat{\mathbf{C}}(\tf r, f^{adv}_\Delta) - \hat{\mathbf{C}}(\tf r, f^{clean}) \big\rVert_2^2 
\end{aligned}
\end{equation}
\vspace{-1.5em}

where $f^{adv}_{\Delta}=f(\tf x, \tf d, \tf e_\Delta)$ and $f^{clean}=f(\tf x, \tf d, \tf e)$ in which $\tf e_\Delta=\{E(\tf I_i + \tfg \delta_i)[\pi_i(\tf x)]\}$.
To solve Eq.~(\ref{eq:rgb_new}), we iteratively update $\tfg \delta_i$ with gradient ascent using an Adam optimizer, where the $t$-th iterative step can be formulated as:

\vspace{-1em}
\begin{equation} \label{eq:update_delta}
    \tfg \delta^{(t+1)}_i = clip(\tfg \delta^{(t)}_i + \eta \cdot Adam (\nabla_{\tfg \delta^{(t)}_i} \tilde{\mathcal{L}}_{rgb}), -\epsilon, \epsilon)
\end{equation}
\vspace{-2em}

where $\eta$ is the learning rate and $clip(\cdot, -\epsilon, \epsilon)$ denotes a clipping operation to constrain the norm of $\tfg \delta_i$.

\textbf{Evaluation setting.} 
\underline{GNeRF variants:} We consider three state-of-the-art (SOTA) GNeRF methods: IBRNet~\cite{wang2021ibrnet}, MVSNeRF~\cite{chen2021mvsnerf}, and GNT~\cite{wang2022attention}, where we adopt their official implementation and load their pretrained models for evaluation. 
\underline{Datasets:} 
We follow the train/test dataset splits adopted by these three GNeRF variants and use both synthetic objects and real scenes from three datasets: three Lambertian objects from DeepVoxels~\cite{sitzmann2019deepvoxels}, eight Realistic Synthetic objects from NeRF~\cite{mildenhall2020nerf}, and eight complex real-world forward-facing scenes from LLFF~\cite{mildenhall2019local}. Regarding the source view selection, we follow each GNeRF variant's default scheme, e.g., select the nearby $N$ views around the target view for IBRNet/GNT.
\underline{NeRFool setup:} The learning rate $\eta$ in Eq.~(\ref{eq:update_delta}) is set to 1e-3 and $\tfg \delta_i$ is initialized with a uniform distribution $\mathcal{U}(-\epsilon, \epsilon)$ and then optimized for 500 iterations.

\begin{table*}[t]
  \vspace{-1em}
  \centering
  \caption{The achieved rendering quality of IBRNet, which is attacked by NeRFool under different numbers of source views and perturbation strength $\epsilon$, on three datasets. The reported results are averaged across all scenes in each dataset. ``Clean" denotes no attack is performed.}
%   \vspace{-0.5em}
    \resizebox{0.9\textwidth}{!}{
\begin{tabular}{cccccccccccc}
\toprule
\multirow{2}{*}{\begin{tabular}[c]{@{}c@{}}\textbf{Attack} \\ \textbf{Method}\end{tabular}} &
  \multirow{2}{*}{\begin{tabular}[c]{@{}c@{}}\textbf{No. Source} \\ \textbf{Views}\end{tabular}} &
  \multirow{2}{*}{\textbf{$\epsilon$}} &
  \multicolumn{3}{c}{\textbf{LLFF}} &
  \multicolumn{3}{c}{\textbf{NeRF Synthetic}} &
  \multicolumn{3}{c}{\textbf{DeepVoxels}} \\ \cmidrule{4-12}
        &    &    & \textbf{PSNR $\uparrow$}  & \textbf{SSIM $\uparrow$} & \textbf{LPIPS $\downarrow$} & \textbf{PSNR $\uparrow$} & \textbf{SSIM $\uparrow$} & \textbf{LPIPS $\downarrow$} & \textbf{PSNR $\uparrow$} & \textbf{SSIM $\uparrow$} & \textbf{LPIPS $\downarrow$} \\\midrule \midrule
\rowcolor{gray}
Clean   & 4  & -  & 23.73 & 0.77 & 0.24  & 28.78 & 0.96 & 0.06  & 32.90 & 0.98 & 0.03  \\ 
NeRFool & 4  & 8  & 13.30 & 0.53 & 0.45  & 12.57 & 0.82 & 0.24  & 12.91 & 0.76 & 0.24  \\
NeRFool & 4  & 16 & 11.99 & 0.45 & 0.51  & 10.71 & 0.75 & 0.31  & 11.85 & 0.71 & 0.29  \\
NeRFool & 4  & 32 & 11.51 & 0.41 & 0.54  & 9.42  & 0.70 & 0.36  & 11.64 & 0.71 & 0.30  \\ \midrule \midrule
\rowcolor{gray}
Clean   & 6  & -  & 24.52 & 0.80 & 0.22  & 29.18 & 0.96 & 0.05  & 34.08 & 0.98 & 0.02  \\
NeRFool & 6  & 8  & 13.00 & 0.57 & 0.42  & 14.58 & 0.86 & 0.21  & 13.49 & 0.81 & 0.21  \\ \midrule \midrule
\rowcolor{gray}
Clean   & 10 & -  & 25.13 & 0.82 & 0.21  & 30.00 & 0.96 & 0.05  & 34.57 & 0.99 & 0.02  \\
NeRFool & 10 & 8  & 12.86 & 0.60 & 0.39  & 10.56 & 0.79 & 0.26  & 11.63 & 0.77 & 0.27  \\ \bottomrule
\end{tabular}
    }
  \vspace{-1em}
  \label{tab:ibrnet_vc}%
\end{table*}%

\subsection{Is GNeRF Robust to Adversarial Perturbations?}
 % \vspace{-0.3em}

\textbf{Attack the most representative GNeRF.} We first apply the above view-specific attach method to IBRNet~\cite{wang2021ibrnet}, which serves as a cornerstone for other GNeRF variants, with varied numbers of source views and perturbation strength $\epsilon$. The corresponding quantitative results and qualitative visualization are shown in Tab.~\ref{tab:ibrnet_vc} and Fig.~\ref{fig:ibrnet_vc}, respectively. We can see that \underline{(1)} our proposed view-specific attack method can considerably degrade the reconstruction accuracy, e.g., a 10.43/11.74 PSNR reduction on average with $\epsilon=$8/16, respectively, on LLFF; \underline{(2)} Imperceptible perturbations, which look like random noise caused by camera shake, in the source views can cause serious unrealistic artifacts in the rendered outputs and thus severally degrade users' visual experience; and \underline{(3)} increasing the number of source views can result in larger PSNR degradation, e.g., a 0.44 larger PSNR reduction when conditioning on ten source views than that of four source views. This indicates that although increased conditionality favors better clean reconstruction accuracy measured on clean source views, it can incur more severe security concerns due to the corresponding higher flexibility (i.e., more pixels) for injecting perturbations. Therefore, in the following experiments, we adopt four source views and $\epsilon=8$ if not specifically stated.

\begin{table}[h]
  \vspace{-0.5em}
  \centering
  \caption{Apply NeRFool on top of other SOTA GNeRF designs.}
%   \vspace{-0.5em}
    \resizebox{0.98\linewidth}{!}{
\begin{tabular}{cccccc}
\toprule
\multirow{2}{*}{\textbf{GNeRF}} & \multirow{2}{*}{\begin{tabular}[c]{@{}c@{}}\textbf{Attack} \\ \textbf{Method}\end{tabular}} & \multirow{2}{*}{\textbf{$\epsilon$}} & \multicolumn{3}{c}{\textbf{LLFF (Avg.)}} \\
                         &         &    & \textbf{PSNR} $\uparrow$ & \textbf{SSIM} $\uparrow$ & \textbf{LPIPS} $\downarrow$ \\ \midrule \midrule
\rowcolor{gray}
\multirow{3}{*}{\begin{tabular}[c]{@{}c@{}}MVS \cellcolor{white}\\ -NeRF \cellcolor{white} \end{tabular}} \cellcolor{white} & Clean   & -  & 23.24 & 0.78 & 0.20  \\ 
                         & NeRFool & 8  & 16.57 & 0.36 & 0.57  \\
                         & NeRFool & 16 & 14.91 & 0.22 & 0.65  \\ \midrule \midrule
                         \rowcolor{gray}
\multirow{3}{*}{GNT} \cellcolor{white}     & Clean   & -  & 23.66 & 0.80 & 0.16  \\
                         & NeRFool & 8  & 14.28 & 0.50 & 0.36  \\
                         & NeRFool & 16 & 12.49 & 0.38 & 0.44  \\ \bottomrule
\end{tabular}
    }
  \label{tab:more_gnerf}%
\vspace{-1.5em}
\end{table}%

\textbf{Attack other GNeRF variants.} We further apply NeRFool's attack method to MVSNeRF~\cite{chen2021mvsnerf} and GNT~\cite{wang2022attention} on the LLFF  dataset and report the average metrics across all test scenes. As shown in Tab.~\ref{tab:more_gnerf}, we can see that \underline{(1)} Our NeRFool attack can consistently degrade the reconstruction accuracy across all GNeRF variants and datasets, e.g., an 8.33/11.17 PSNR reduction for MVSNeRF/GNT, respectively, when $\epsilon$=16;
\underline{(2)} GNT can improve the adversarial robustness over IBRNet with reduced PNSR degradation, maybe because of the former's newly introduced transformer modules, which increase the non-linearity of the overall GNeRF pipeline; \underline{(3)} MVSNeRF wins the highest level of robustness among the three GNeRF variants thanks to its accurate geometry estimation~\cite{chen2021mvsnerf}, which is of great significance for GNeRF's robustness as analyzed in Sec.~\ref{sec:density_or_color}.

\textbf{Key insight.} Although GNeRF involves more diverse operations than DNNs, adversarially perturbing its 2D source views can still considerably destruct its reconstructed 3D scene rendered from specific views. Furthermore, while increased conditionality on source views in GNeRF can boost its cross-scene generalization capability, it comes at the cost of higher security concerns.

\begin{table*}[t]
  \vspace{-1em}
  \centering
  \caption{Apply NeRFool on top of IBRNet and MVSNeRF on LLFF via perturbing the color, density, or both. \cmark marks the perturbed item.}
%   \vspace{-0.5em}
    \resizebox{0.98\linewidth}{!}{
\begin{tabular}{c|cccc|cccccccc}
\toprule
\multirow{2}{*}{\textbf{Method}} &
\multicolumn{2}{c}{\textbf{Color}}                   & \multicolumn{2}{c|}{\textbf{Density}}                 & \multicolumn{8}{c}{\textbf{Achieved PSNR $\uparrow$} }                                            \\ \cmidrule{2-13}
& \textbf{Clean}          & \textbf{Adv.}                  & \textbf{Clean}                & \textbf{Adv.}                  & \textbf{fern}  & \textbf{flowers} & \textbf{fortress} & \textbf{horns} & \textbf{leaves} & \textbf{orchids} & \textbf{room}  & \textbf{trex}  \\ \midrule
\rowcolor{gray}
\multirow{6}{*}{\textbf{IBRNet}} \cellcolor{white} &
\cmark            & \multicolumn{1}{c}{} & \cmark  & \multicolumn{1}{c|}{}          &  22.22 & 25.93 & 28.42 & 24.39 & 18.93 & 18.35 & 28.68 & 22.91 \\ \cmidrule{2-13}
& \cmark            & \multicolumn{1}{c}{} & \multicolumn{1}{c}{} & \cmark            & 15.77 & 18.10   & 16.02    & 15.07 & 15.54  & 12.72   & 17.56 & 14.19 \\ \cmidrule{2-13}
& \multicolumn{1}{l}{} & \cmark         & \cmark          & \multicolumn{1}{l|}{} & 21.77 & 23.12   & 27.99    & 23.15 & 23.15  & 16.96   & 26.74 & 22.13 \\ \cmidrule{2-13}
 & \multicolumn{1}{l}{} & \cmark            & \multicolumn{1}{l}{} & \cmark         & 14.02 \cellcolor{highlight}  & 15.10 \cellcolor{highlight}   & 13.46 \cellcolor{highlight}    & 12.41 \cellcolor{highlight} & 14.57 \cellcolor{highlight} & 11.54 \cellcolor{highlight}   & 12.96 \cellcolor{highlight} & 12.31 \cellcolor{highlight} \\ \midrule
\rowcolor{gray}
 \multirow{6}{*}{\textbf{MVSNeRF}} \cellcolor{white} &
\cmark            & \multicolumn{1}{c}{} & \cmark  & \multicolumn{1}{c|}{}          &  22.10 & 25.52 & 28.21 & 23.87 & 18.12 & 17.92 & 28.13 & 22.03 \\ \cmidrule{2-13}
& \cmark            & \multicolumn{1}{c}{} & \multicolumn{1}{c}{} & \cmark            & 17.87 & 20.93   & 17.22    & 17.48 & 14.36  & 15.73   & 15.85 \cellcolor{highlight} & 16.21 \\ \cmidrule{2-13}
& \multicolumn{1}{l}{} & \cmark         & \cmark          & \multicolumn{1}{l|}{} & 17.75 & 20.14  & 15.44 \cellcolor{highlight}   & 18.51 & 14.45  & 15.57  & 19.75 & 18.36 \\ \cmidrule{2-13}
 & \multicolumn{1}{l}{} & \cmark            & \multicolumn{1}{l}{} & \cmark         & 17.13 \cellcolor{highlight}& 19.63\cellcolor{highlight}  & 15.95   & \cellcolor{highlight} 17.04 & \cellcolor{highlight} 13.96  & \cellcolor{highlight} 15.31  & 17.65 & \cellcolor{highlight} 15.96 \\ \bottomrule
\end{tabular}
    }
  \vspace{-1em}
  \label{tab:exp_perturb_one}%
\end{table*}%

\subsection{What to Perturb: Density, Color, or Both?}
\label{sec:density_or_color}
 % \vspace{-0.3em}

Considering that NeRF's rendered pixels are alpha-composited from both estimated density and color, one natural question regarding GNeRF's vulnerability is ``\textit{which component is easier to be perturbed by adversarial perturbations, density, color, or both}"? We aim to answer this question with the following experiments and discussions.

\begin{figure}[t]
\vspace{-1em}
\centering
\includegraphics[width=0.95\linewidth]{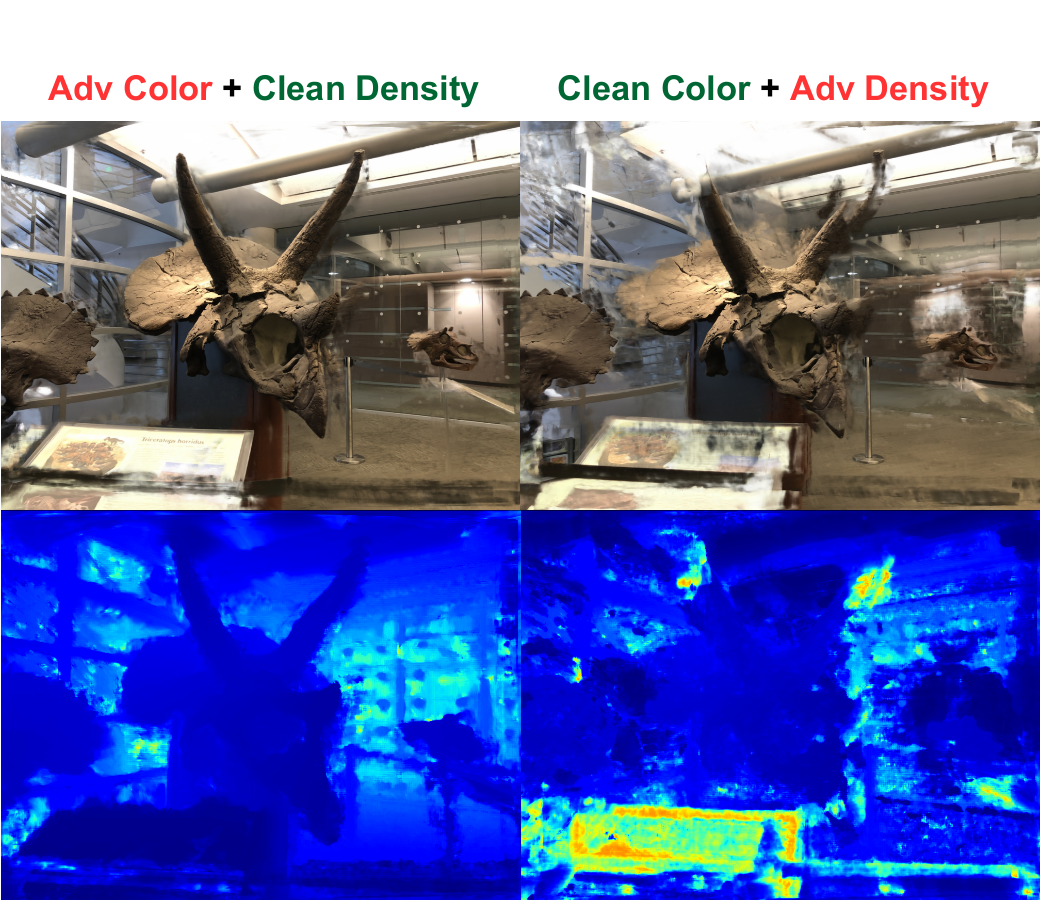}
\vspace{-1em}
\caption{Visualize the attack effectiveness of NeRFool on IBRNet on top of LLFF achieved by perturbing either color or density.}
\label{fig:dens_or_color}
\vspace{-0.5em}
\end{figure}

\begin{figure}[t]
% \vspace{-1em}
\centering
\includegraphics[width=0.98\linewidth]{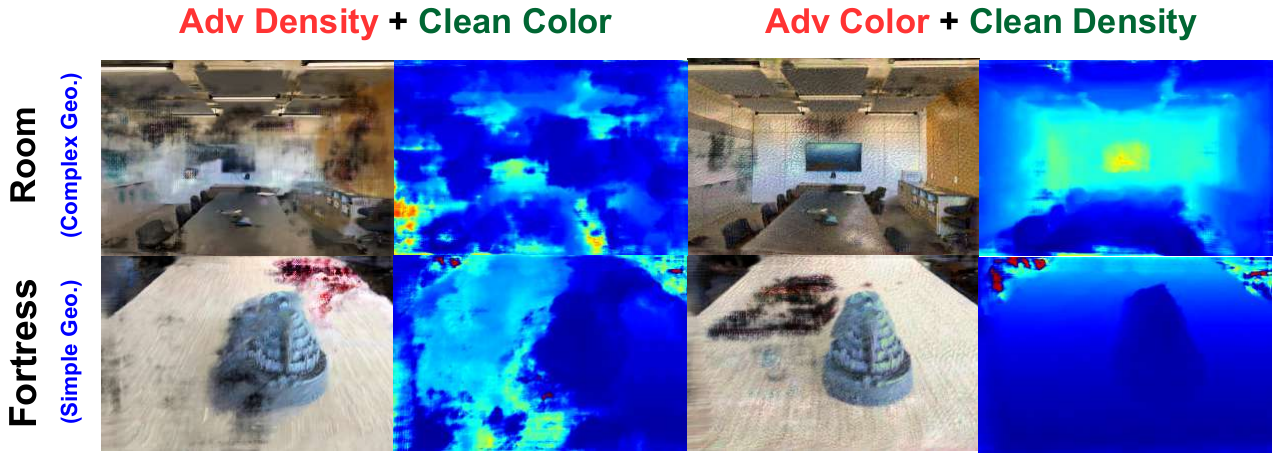}
\vspace{-1em}
\caption{Attack MVSNeRF by perturbing either color or density on two scenes with complex/simple geometry, respectively.}
\label{fig:mvsnerf_perturb_one}
\vspace{-1.5em}
\end{figure}

\textbf{Setup.} We conduct an ablation study on top of IBRNet~\cite{wang2021ibrnet} and MVSNeRF~\cite{chen2021mvsnerf}, in which we perform the alpha-composition via (a) clean densities $\sigma_{clean}$ plus perturbed colors $\tf c_{adv}$, and (b) perturbed densities $\sigma_{adv}$ plus clean colors $\tf c_{clean}$. To implement this setting, we first acquire $(\sigma_{clean}, \tf c_{clean})$ on top of clean source views and $(\sigma_{adv}, \tf c_{adv})$ on top of perturbed source views, and next conduct volume rendering using (a) $(\sigma_{clean}, \tf c_{adv})$ and (b) $(\sigma_{adv}, \tf c_{clean})$, respectively, based on Eq.~(\ref{eq:rendering}).
We summarize the achieved PSNR in Tab.~\ref{tab:exp_perturb_one} and visualize both the rendered RGB images and the estimated depth, the latter of which is derived by replacing $\mathbf{c}(t)$ in Eq.~(\ref{eq:rendering}) with the ray depth $t$, in Fig.~\ref{fig:dens_or_color} and Fig.~\ref{fig:mvsnerf_perturb_one}.

\textbf{Observation.} As shown in Tab.~\ref{tab:exp_perturb_one}, we can observe that \underline{(1)} 
using $(\sigma_{adv}, \tf c_{clean})$ can already considerably degrade the rendering quality, resulting in a comparable PSNR degradation as compared to perturbing both the densities and colors.
As further verified in Fig.~\ref{fig:dens_or_color}, using $(\sigma_{adv}, \tf c_{clean})$ on top of IBRNet can considerably destruct both the rendered RGB images and depth, where the regions with unrealistic depth estimation cause more severe artifacts in the corresponding RGB images; \underline{(2)}  $(\sigma_{clean}, \tf c_{adv})$ shows poor attack effectiveness on IBRNet, causing very limited PSNR degradation; 
\underline{(3)} $(\sigma_{adv}, \tf c_{clean})$ can achieve a 0.62 larger PSNR reduction on average over $(\sigma_{clean}, \tf c_{adv})$ on MVSNeRF and their 
rankings of attack effectiveness vary across scenes. In particular, as shown in Fig.~\ref{fig:mvsnerf_perturb_one}, $(\sigma_{adv}, \tf c_{clean})$ can result in larger PSNR degradation on scenes with more complex geometry (e.g., the indoor scene ``room" in LLFF), where the depth of different objects in a scene varies significantly and thus correctly rendering the RGB images relies more on accurate geometry estimation. Instead, on scenes with relatively simpler geometry (e.g., the ``fortress" composed of a table and an object in LLFF), perturbing colors can win better attack effectiveness; \underline{(4)} perturbing both densities and colors can lead to larger PSNR degradation as compared to only perturbing one factor in 14 out of 16 cases.

\textbf{Key insight.} 
This set of experiments indicates that \underline{(1)} adversarial perturbations tend to be more effective in perturbing the density than perturbing the color, especially for  scenes with complex geometries, which we conjecture is because the induced wrong geometry estimation of the former can more severely degrade the reconstructed images. This insight could inspire novel attacks (e.g., our NeRFool$^+$ in Sec.~\ref{sec:nerfool}) and defense methods dedicated to GNeRF;
\underline{(2)} we conjecture that the stronger robustness of IBRNet against perturbed colors over MVSNeRF may originate from a more robust color prediction scheme. In particular, to derive the color of a sampled point, instead of directly regressing the RGB value as in MVSNeRF~\cite{chen2021mvsnerf}, IBRNet projects it to all source views and predicts the weights for blending the RGB values of its projection points on different source views~\cite{wang2021ibrnet}, which could result in marginal color perturbations when the projection points share similar RGB values. This insight could shed light on the design of more robust GNeRF pipelines.

\begin{table}[h]
  \vspace{-1em}
  \centering
  \caption{Apply NeRFool on IBRNet w/ and w/o per-scene finetuning. The achieved PSNR on each scene is reported.}
%   \vspace{-0.5em}
    \resizebox{\linewidth}{!}{
\begin{tabular}{cc|cccc}
\toprule
\multicolumn{2}{c|}{\textbf{Scenes}}       & \textbf{fern}  & \textbf{flower} & \textbf{fortress} & \textbf{horns} \\ \midrule
\rowcolor{gray}
\multirow{2}{*}{w/o ft.} \cellcolor{white} & clean \cellcolor{white} & 22.22 & 25.93  & 28.42    & 24.39 \\
                         & adv.   & 14.02 & 15.10   & 13.46    & 12.41  \\ \midrule \midrule
\rowcolor{gray}
\multirow{2}{*}{w/ ft.} \cellcolor{white}  & clean \cellcolor{white} & 24.10 & 27.13  & 30.64    & 27.83 \\
                         & adv.   & 13.93 & 12.29  & 12.71     & 11.78 \\ \bottomrule
\end{tabular}
}
  \vspace{-1em}
  \label{tab:per_scene_ft}%
\end{table}%

\subsection{How Per-Scene Finetuning Impacts  Robustness?}
\label{sec:finetuning}
 % \vspace{-0.3em}

While per-scene finetuning can be adopted on top of GNeRF to enhance the reconstruction accuracy~\cite{wang2021ibrnet,chen2021mvsnerf,xu2022point,liu2022neural}, its implication on GNeRF's adversarial robustness is unknown. Here we study the robustness of finetuned GNeRF on different scenes using IBRNet~\cite{wang2022attention}.

\textbf{Observation.} Tab.~\ref{tab:per_scene_ft} shows that while finetuning can boost the clean reconstruction accuracy, the accuracy degradation caused by adversarial perturbations becomes worse.

\textbf{Key insight.}
We conjecture this is because the resulting density and color from per-scene finetuned GNeRF are more overfitted to each scene, which could harm the model robustness according to the previous robustness insights for DNNs~\cite{rice2020overfitting}. This finding calls for robust per-scene finetuning schemes dedicated to GNeRF to reduce overfitting and maximize robustness.

\begin{table}[t]
  \vspace{-0.5em}
  \centering
  \caption{Benchmark view-specific attacks and transferred attacks.}
%   \vspace{-0.5em}
    \resizebox{0.98\linewidth}{!}{
\begin{tabular}{ccccc}
\toprule
\multirow{2}{*}{\textbf{GNeRF}} & \multirow{2}{*}{\begin{tabular}[c]{@{}c@{}}\textbf{Attack}\\ \textbf{Mode}\end{tabular}} & \multicolumn{3}{c}{\textbf{LLFF (Avg.)}} \\ 
                        &               & \textbf{PSNR $\uparrow$} & \textbf{SSIM $\uparrow$} & \textbf{LPIPS $\downarrow$} \\ \midrule \midrule
\rowcolor{gray}
\multirow{3}{*}{IBRNet} \cellcolor{white} & Clean         & 23.73 & 0.77 & 0.24  \\
                        & Transfer      & 23.36 & 0.76 & 0.27  \\
                        & View-specific & 13.30 & 0.53 & 0.45  \\ \midrule \midrule
\rowcolor{gray}
\multirow{3}{*}{GNT}  \cellcolor{white}  & Clean         & 23.66 & 0.16 & 0.80  \\
                        & Transfer      & 21.92 & 0.73 & 0.20  \\
                        & View-specific & 14.28 & 0.36 & 0.50  \\ \bottomrule
\end{tabular}
}
  \vspace{-1.5em}
  \label{tab:tranferability}%
\end{table}%

\begin{figure*}[t!]
\centering
\includegraphics[width=0.8\linewidth]{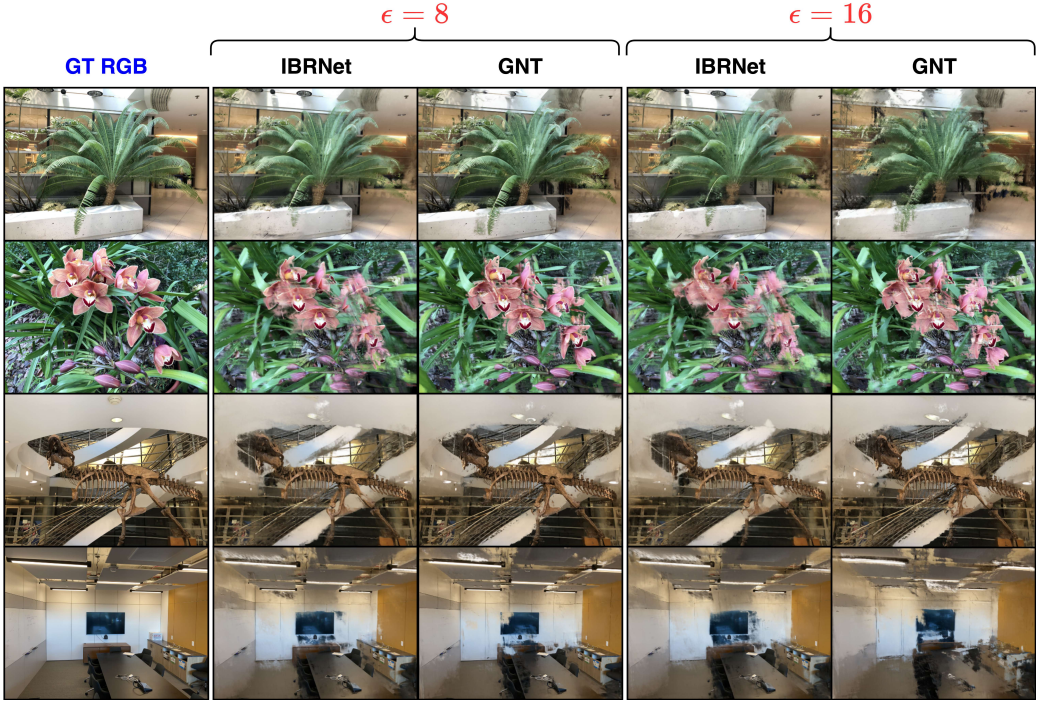}
\vspace{-1em}
\caption{Visualize the attack effectiveness of NeRFool$^+$ on IBRNet/GNT on the scenes from LLFF under different perturbation strengths.}
\label{fig:universal}
\vspace{-0.5em}
\end{figure*}

 % \vspace{-0.5em}
\subsection{Are the Perturbations Transferable across Views?}
\label{sec:transferability}

In real-world settings, it is more practical and highly desirable to reuse the same adversarial perturbations to fool a wide range of target views, under which only the perturbed source views need to be provided to users for conducting effective attacks. To achieve this, it requires that the generated adversarial perturbations can transfer (i.e., remain effective) across target views, motivating the following study.

\textbf{Setup.} 
We optimize $\tfg \Delta$ in 
Eq.~(\ref{eq:objective}) based on a sampled target view centering around each test scene and then reuse it to fool other target views on top of IBRNet.

\textbf{Observation.} Tab.~\ref{tab:tranferability} shows that under a transferring setting, the achieved PSNR degradation is considerably lower than that achieved by the above view-specific attack.

\textbf{Key insight.} This indicates that $\tfg\Delta$ optimized for one specific view can be hardly transferred across different views. We understand that this is because camera rays emitted from one target view can only cover a limited set of 3D points in a scene and thus it is difficult for $\tfg\Delta$ optimized for these 3D points to fool other 3D points along the rays emitted from other views with different camera poses.

\section{NeRFool$^+$: Towards Universal Adversarial Perturbations Across Different Views}
\label{sec:nerfool}
 % \vspace{-0.3em}
 
As analyzed in Sec.~\ref{sec:transferability}, generating universal adversarial perturbations that are transferable across different target views can better uncover GNeRF's vulnerability under a more practical setting and enhance our understanding of their deployability in real-world applications. To achieve this, we propose NeRFool$^+$ which integrates two across-view attack techniques dedicated to GNeRF.

 % \vspace{-0.5em}
\subsection{Overview}
 % \vspace{-0.3em}

\textbf{Inspirations from the above exploration.} 
Our NeRFool$^+$'s attack techniques are inspired by the following two insights: \underline{(1)} to enhance the transferability of $\tfg\Delta$, more 3D points on the rays of different views for the target 3D scene, are needed to be sampled, according to the analysis in Sec.~\ref{sec:transferability}; and \underline{(2)} considering that adversarial perturbations are more effective in perturbing the density/geometry based on our findings in Sec.~\ref{sec:density_or_color}, leveraging estimated geometry as extra supervision could enhance the optimization on $\tfg \Delta$.

\textbf{Two optimization techniques.} Leveraging the aforementioned insights, our NeRFool$^+$ integrates two optimization techniques correspondingly: \underline{(1)} to cover more rays and sample more 3D points in a scene, we sample unseen novel target views via spherical linear interpolation among known camera poses to augment the training sets; \underline{(2)} to better ruin the geometry prediction, we maximize the depth estimation error as extra supervision when optimizing $\tfg \Delta$. The technical details are elaborated below.

 % \vspace{-0.5em}
\subsection{Novel Target View Sampling}
 % \vspace{-0.3em}
\textbf{Sampling strategy.}
One intuitive sampling strategy is to randomly sample along the upper hemisphere of the target scene. However, this may not be applicable to new scenes where a sampling boundary is hard to define. To develop a sampling strategy generally applicable to new scenes with arbitrary view distributions, we instead randomly interpolate the known camera poses of the source views (and training views if available in the dataset) to create novel views during each training iteration. One advantage is that such a sampling scheme can implicitly define a meaningful range of possible camera poses.

\textbf{Interpolation strategy.}
In NeRFool$^+$, given two randomly selected known camera poses $\tf P_{1}=[\tf R_{1} \mid \tf T_{1} ]$ and 
$\tf P_{2}=[\tf R_{2} \mid \tf T_{2} ]$, we interpolate both their rotation matrices and translation vectors to acquire new ones $\tf P_{new}=[\tf R_{new} \mid \tf T_{new} ]$.
In particular, a linear interpolation is performed for the translation vectors: $\tf T_{new} = \alpha \tf T_{1} + (1-\alpha) \tf T_{2}$, where $\alpha \sim \mathcal{U}(0, 1)$. For ensuring meaningful $\tf R_{new}$ on a unit-radius great circle, we adopt spherical linear interpolation (Slerp)~\cite{shoemake1985animating}: $\tf R_{new}=Slerp(\tf R_{1}, \tf R_{2}; \alpha)$, following the formulation in~\cite{shoemake1985animating}.
The pseudo-RGB ground truth is then reconstructed for $\tf P_{new}$ as in Sec.~\ref{sec:nerfool-} to optimize $\tfg \Delta$.

 % \vspace{-0.5em}
\subsection{Geometric Error Maximization}
 % \vspace{-0.3em}

To better ruin the geometry prediction via extra supervision, we maximize the depth estimation error under the guidance of a pretrained depth estimation model $F_{D}$, following~\cite{xu2022sinnerf}, which is used to provide geometry priors. In particular, we apply $F_{D}$ on the reconstructed pseudo-RGB ground truth to generate the depth map as supervision signals, where the objective can be formulated as:

\vspace{-0.5em}
\begin{equation}
    \mathcal{L}_{depth} = \sum_{\tf r \in \mathcal{R}} \left\lVert
   \hat{\mathbf{D}}(\tf r, f^{adv}_{\tfg \Delta}) - F_{D}(\hat{\mathbf{C}}(\tf r, f^{clean})) \right\rVert_2^2
    \label{eq:depth_error}
\end{equation}
\vspace{-1em}

where $\hat{\mathbf{D}}(\mathbf{r}, f) = \int_{t_n}^{t_f} T(t) \sigma(\tf r(t)) t \dif t$
and $\mathcal{R}$ are sampled from interpolated $\tf P_{new}$ during each iteration.

\begin{figure}[t]
% \vspace{-1em}
\centering
\includegraphics[width=0.98\linewidth]{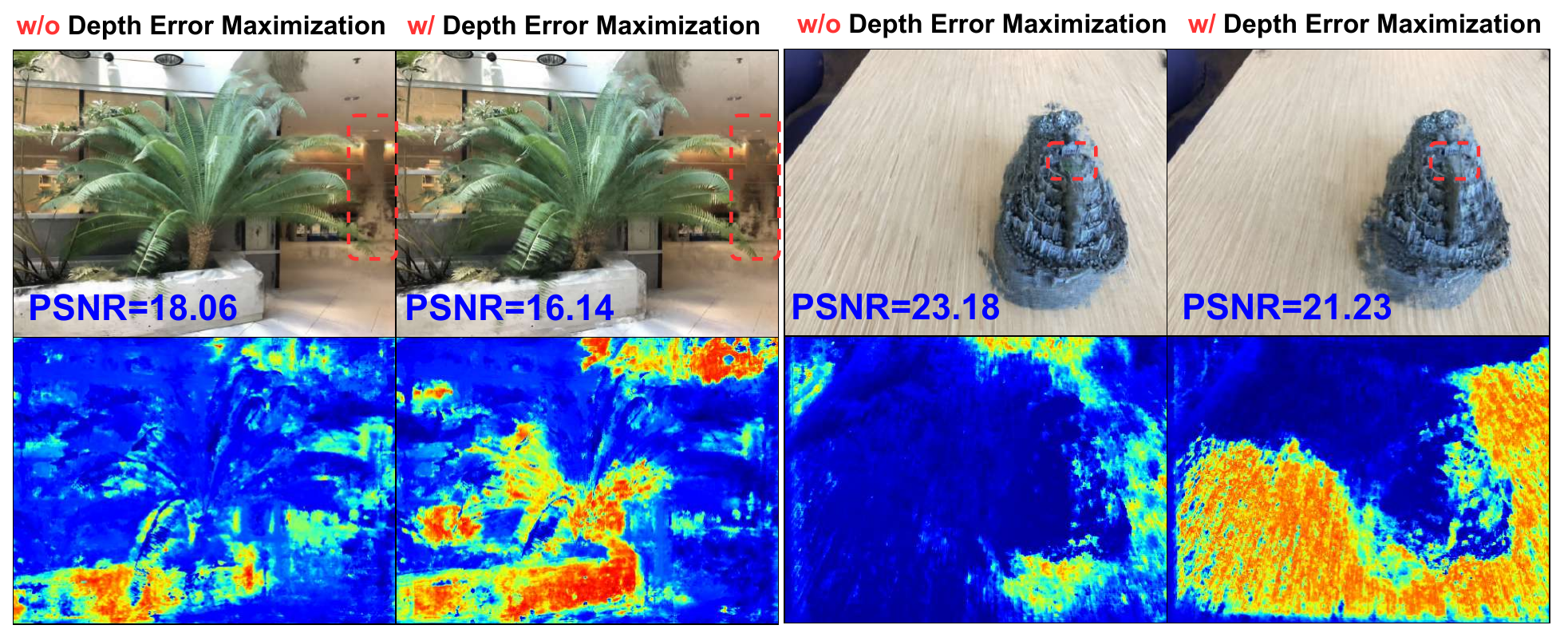}
\vspace{-0.5em}
\caption{Apply NeRFool$^+$ on LLFF w/ and w/o $\mathcal{L}_{depth}$.}
\label{fig:depth_ablation}
\vspace{-1em}
\end{figure}

 % \vspace{-0.5em}
\subsection{Evaluating NeRFool$^+$}
 % \vspace{-0.3em}

\textbf{Setup.} We adopt the same evaluation as in Sec.~\ref{sec:nerfool-}, except that we adopt the same set of source views, which are randomly sampled from nearby views of the target forward-facing scene and then fixed for all runs, for all target view directions. We reuse the pretrained depth estimation model $F_{D}$ in~\cite{xu2022sinnerf}. We adopt four source views and $\epsilon=8$ by default if not specifically stated.

\textbf{Observation and analysis.}
As shown in Tab.~\ref{tab:nerfool} and Fig.~\ref{fig:universal}, we can observe that \underline{(1)} our NeRFool$^+$ can consistently introduce severe artifacts in the reconstructed target views across different GNeRF variants, and considerably degrade the reconstruction PSNR, e.g., an up-to 17.26 lower PSNR on GNT under $\epsilon$=16, as compared to the clean results; \underline{(2)} our NeRFool$^+$ can induce significantly larger PSNR degradation as compared to the transferred perturbations via NeRFool in Tab.~\ref{tab:tranferability}, indicating the effectiveness of NeRFool$^+$ in enhancing the attack transferability across target views.

\begin{table}[t]
  \vspace{-0.5em}
  \centering
  \caption{The achieved attack effectiveness of NeRFool$^+$ on three GNeRF designs on the scenes from LLFF under different perturbation strengths $\epsilon$. The PSNR on each scene is reported.}
%   \vspace{-0.5em}
    \resizebox{\linewidth}{!}{
\begin{tabular}{cccccccccc}
\toprule
\textbf{GNeRF} & \textbf{$\epsilon$}  & \textbf{fern} & \textbf{flower} & \textbf{fortress} & \textbf{horns} & \textbf{leaves} & \textbf{orchids} & \textbf{room} & \textbf{trex} \\ \midrule
\rowcolor{gray}
\multirow{3}{*}{IBRNet} \cellcolor{white} & 0  & 22.22 & 25.93 & 28.42 & 24.39 & 18.93 & 18.35 & 28.68 & 22.91 \\ 
 & 8  & 17.13 & 14.73 & 12.04 & 12.70  & 13.05 & 12.48 & 13.60  & 13.79 \\
                         & 16  & 15.97 & 13.69 & 11.45 & 12.04 & 12.29 & 11.51 & 12.77 & 13.23 \\
                         \midrule \midrule
\rowcolor{gray}
\multirow{3}{*}{GNT}  \cellcolor{white}   & 0  & 22.58 & 24.93 & 29.08 & 25.02 & 18.8  & 17.69 & 28.20  & 22.96 \\ 
                         & 8  & 17.45 & 15.20  & 12.81 & 13.50  & 13.06 & 13.10  & 14.86 & 13.78 \\
                         & 16 & 13.38 & 11.48 & 11.82 & 11.91 & 10.32 & 11.62 & 11.96 & 12.94 \\ 
                         \midrule \midrule
\rowcolor{gray}
\multirow{3}{*}{MVSNeRF} \cellcolor{white} & 0  & 22.10  & 25.52 & 28.21 & 23.87 & 18.12 & 17.92 & 28.13 & 22.03 \\ 
                         & 8  & 16.18 & 19.84 & 12.26 & 17.17 & 13.51 & 15.58 & 17.78 & 16.52 \\
                         & 16 & 15.15 & 17.19 & 12.05 & 14.23 & 12.84 & 14.60 & 15.99 & 13.60  \\
                         \bottomrule
\end{tabular}
}
  \vspace{-0.5em}
  \label{tab:nerfool}%
\end{table}%

\textbf{The role of $\mathcal{L}_{depth}$.}
We conduct an ablation study for NeRFool$^+$ w/ and w/o enabling $\mathcal{L}_{depth}$. As shown in Fig.~\ref{fig:depth_ablation}, we find that \underline{(1)} for both NeRFool$^+$ w/ and w/o $\mathcal{L}_{depth}$, the target views farther from the scene, i.e., on the sampling boundary, are harder to be destructed, since 3D points along their emitted rays are less likely to be sampled as compared to those from the center views; and \underline{(2)} NeRFool$^+$ w/ $\mathcal{L}_{depth}$ can cause larger artifacts, e.g., a 1.92 larger PSNR degradation on fern over NeRFool$^+$ w/o $\mathcal{L}_{depth}$, on the aforementioned target views farther from the scene thanks to the supervision from scene geometry priors.

\begin{table}[t]
  \vspace{-0.5em}
  \centering
  \caption{Visualize the achieved PSNR of GNT~\cite{wang2022attention} under different pretraining and test scheme combinations. ``Pre." denotes the pertaining scheme.}
%  \vspace{-0.5em}
    \resizebox{0.98\linewidth}{!}{
\begin{tabular}{cc|cccc}
\toprule
\textbf{Pre.} & \textbf{Test} & \textbf{fern} & \textbf{flower} & \textbf{fortress} & \textbf{horns} \\ \midrule
\rowcolor{gray}
clean & clean & 22.58 & 24.93 & 29.08 & 25.02 \\
clean & adv. & 15.31 & 15.90 & 14.62 & 14.62 \\ \midrule \midrule
adv. & clean & 22.18 & 22.75 & 28.14 & 24.64 \\
adv. & adv. & 20.85 & 19.58 & 25.01 & 22.71 \\ \bottomrule
\end{tabular}
}
  \vspace{-1em}
  \label{tab:adv_pretrain}%
\end{table}%

 \begin{table*}[t]
  \vspace{-1em}
  \centering
  \caption{Visualize the achieved PSNR of IBRNet and GNT under different finetuning and test scheme combinations. ``Ft." denotes the adopted finetuning scheme. The achieved highest robust/clean PSNR on each scene across all settings is highlighted.}
%   \vspace{-0.5em}
    \resizebox{0.99\linewidth}{!}{
\begin{tabular}{c|cccc|cccccccc}
\toprule
\multirow{2}{*}{\textbf{Method}} &
\multicolumn{2}{c}{\textbf{Ft.}}                   & \multicolumn{2}{c|}{\textbf{Test}}                 & \multicolumn{8}{c}{\textbf{Achieved PSNR $\uparrow$} }                                            \\ \cmidrule{2-13}
& \textbf{Clean}          & \textbf{Adv.}                  & \textbf{Clean}                & \textbf{Adv.}                  & \textbf{fern}  & \textbf{flowers} & \textbf{fortress} & \textbf{horns} & \textbf{leaves} & \textbf{orchids} & \textbf{room}  & \textbf{trex}  \\ \midrule
\rowcolor{gray}
\multirow{6}{*}{\textbf{IBRNet}} \cellcolor{white} &
\cmark            & \multicolumn{1}{c}{} & \cmark  & \multicolumn{1}{c|}{}          &  24.10 & 27.13 & 30.64 & 27.83 & 21.12 & 20.32 & 31.45 & 25.77  \\ \cmidrule{2-13}
& \cmark            & \multicolumn{1}{c}{} & \multicolumn{1}{c}{} & \cmark            & 13.93 & 12.29 & 12.71 & 11.78 & 13.25 & 10.35 & 12.87 & 13.26 \\ \cmidrule{2-13}
& \multicolumn{1}{l}{} & \cmark         & \cmark          & \multicolumn{1}{l|}{} & 23.78 & 27.11 & 30.01 & 27.37 & 20.97 & 20.16 & 31.13 & 25.33 \\ \cmidrule{2-13}
 & \multicolumn{1}{l}{} & \cmark            & \multicolumn{1}{l}{} & \cmark         & 23.57 & 26.91 & 29.29 & 26.95 & 20.78 & 19.91 & 30.67 & 25.06 \\ \midrule
\rowcolor{gray}
 \multirow{6}{*}{\textbf{GNT}} \cellcolor{white} &
\cmark            & \multicolumn{1}{c}{} & \cmark  & \multicolumn{1}{c|}{}          &  24.09 & 29.97 & 33.24 & 26.82 & 23.49 & 20.13 & 33.56 & 27.53 \\ \cmidrule{2-13}
& \cmark            & \multicolumn{1}{c}{} & \multicolumn{1}{c}{} & \cmark            & 14.51 & 13.91 & 13.31 & 11.66 & 10.84 & 11.98 & 12.56 & 10.41 \\ \cmidrule{2-13}
& \multicolumn{1}{l}{} & \cmark         & \cmark          & \multicolumn{1}{l|}{} & \cellcolor{highlight} 24.93 & \cellcolor{highlight} 30.84 & \cellcolor{highlight} 33.67 & \cellcolor{highlight} 27.36 & \cellcolor{highlight} 24.01 & \cellcolor{highlight} 20.61 & \cellcolor{highlight} 32.73 & \cellcolor{highlight} 27.82  \\ \cmidrule{2-13}
 & \multicolumn{1}{l}{} & \cmark            & \multicolumn{1}{l}{} & \cmark         & \cellcolor{highlight} 24.19 & \cellcolor{highlight} 27.91 & \cellcolor{highlight} 32.94 & \cellcolor{highlight} 26.97 & \cellcolor{highlight} 23.41 & \cellcolor{highlight} 19.97 & \cellcolor{highlight} 32.59 & 27.46 \cellcolor{highlight}\\ \bottomrule
\end{tabular}
    }
  \vspace{-1em}
  \label{tab:adv_ft}%
\end{table*}%

\section{Defend against NeRFool: Adversarial GNeRF Training}
\label{sec:defense}

Based on the delivered insights from Sec.~\ref{sec:exploration} and Sec.~\ref{sec:nerfool}, we further perform an intriguing investigation on defending against our NeRFool attack as another crucial piece for understanding GNeRF's robustness.

\subsection{Adversarial GNeRF Training: Formulation}

We robustify GNeRF via integrating adversarial training~\cite{goodfellow2014explaining,shafahi2019adversarial,wong2019fast,madry2017towards}, which augments the training scenes with adversarially perturbed source views based on the following formulation: 

\vspace{-0.5em}
\begin{equation} \label{eq:adv_training}
\min_{\tfg \theta} \max_{\forall \tfg \delta_i \in \tfg\Delta: \, \| \tfg \delta_i\|_{\infty} \leq \epsilon} \tilde{\mathcal{L}}_{rgb}(\mathcal{R}_{target}, f_{\tfg \theta}, \tfg\Delta)
\end{equation} 
\vspace{-1.5em}

where $\tfg \theta$ is the weight of GNeRF and the inner optimization on $\tfg\delta_i\in\tfg\Delta$ is performed using PGD~\cite{madry2017towards}. We apply adversarial training to either GNeRF's pretraining or finetuning stages and evaluate the achieved robustness against NeRFool in the following sections.

\subsection{Evaluation: Adversarial Pretraining}
\label{sec:adv_pre}

\textbf{Setup.} We apply the aforementioned adversarial training to GNT's pretraining stage~\cite{wang2022attention} using $\tfg \epsilon$=8 and an iteration of 1 for updating $\tfg \delta_i$. We then evaluate the resulting models'  robustness against NeRFool with four adversarially perturbed source views and $\tfg \epsilon=8$.

\textbf{Observation and analysis.} As shown in Tab.~\ref{tab:adv_pretrain}, we can observe that 
\underline{(1)} adversarial pretraining can effectively boost the adversarial robustness against our NeRFool, e.g., a 10.39 higher PSNR on the scene fortress; 
\underline{(2)}
the boosted robustness comes at the cost of reduced clean PSNR, e.g., a 0.40$\sim$2.18 PSNR reduction across all scenes, which aligns with previous findings in the literature on adversarial robustness~\cite{zhang2019theoretically}. We also note that this set of experiments represents a first-step exploration towards adversarial GNeRF training. Promising future directions include the development of more advanced GNeRF pipelines that can win both accuracy and robustness.

\subsection{Evaluation: Adversarial Finetuning}

\textbf{Setup.} We apply adversarial training to the finetuning stage of pretrained IBRNet~\cite{wang2021ibrnet} and GNT~\cite{wang2022attention} and evaluate the resulting models' robustness against our NeRFool using the same settings in Sec.~\ref{sec:adv_pre}.

\textbf{Observation.} 
As shown in Tab.~\ref{tab:adv_ft}, we can observe that \underline{(1)} adversarial finetuning can more effectively boost the adversarial robustness as compared to adversarial pretraining reported in Tab.~\ref{tab:adv_pretrain}, e.g., a 5.20 PSNR improvement averaged over four scenes on top of GNT; \underline{(2)} adversarial finetuning can maintain a comparable clean PSNR on IBRNet and consistently boost the clean PSNR on GNT across all scenes, e.g., a 0.87 PSNR improvement on the scene flowers.

\textbf{Key insight.} This set of experiments indicates the benign impact of adversarial perturbations beyond robustness. We conjecture that this is because adversarial perturbations can serve as data augmentation to reduce  overfitting during finetuning as observed in Sec.~\ref{sec:finetuning} and thus boost reconstruction accuracy. This aligns with the observations in image classification tasks that properly induced adversarial robustness could boost accuracy~\cite{xie2020adversarial,deng2021adversarial,salman2020adversarially}. This finding highlights the potential of applying verified training techniques from well-studied image classification tasks to enhance GNeRF optimization.
 \vspace{-2em}
\section{Conclusion}
\label{sec:conclusion}
 % \vspace{-0.3em}

GNeRF has gained increasing attention thanks to its potential in enabling instant and real-time rendering of new scenes, whereas 
its adversarial robustness has not yet been studied and understood, which can limit its real-world deployment. Our work is the first to uncover and study the adversarial vulnerability of GNeRF. In particular, our proposed NeRFool framework presents systematic analysis and experiments of various GNeRF variants and discovers important insights regarding GNeRF's adversarial robustness. Furthermore, we develop NeRFool$^+$ to effectively attack GNeRF across a wide range of target views and provide rich insights for defending against our developed attacks.
Our work has opened a new perspective in the  literature of NeRF and could shed light on more robust GNeRF pipelines to empower their real-world deployment.

\vspace{-0.5em}
\section*{Acknowledgement}
This work was supported in part by CoCoSys, one of the seven centers in JUMP 2.0, a Semiconductor Research Corporation (SRC) program sponsored by DARPA.

\bibliography{ref}
\bibliographystyle{icml2023}

\end{document}